\documentclass[11pt]{article}

\usepackage[letterpaper,margin=1in]{geometry}
\usepackage[T1]{fontenc}
\usepackage{lmodern}
\usepackage{microtype}
\usepackage{booktabs}
\usepackage{graphicx}
\usepackage{hyperref}
\usepackage{xcolor}
\usepackage{amsmath}
\usepackage{amssymb}
\usepackage{listings}
\usepackage{caption}
\usepackage{subcaption}

\hypersetup{
  colorlinks=true,
  linkcolor=blue!60!black,
  citecolor=blue!60!black,
  urlcolor=blue!60!black,
  breaklinks=true,
}

\title{\textsc{ShapeCodeBench}: A Renewable Benchmark for Perception-to-Program Reconstruction of Synthetic Shape Scenes%
  \thanks{Code, benchmark data, and figures: \url{https://github.com/shivamk3r/shape-code-bench}. Archived release DOI: \href{https://doi.org/10.5281/zenodo.20132286}{10.5281/zenodo.20132286}.}}

\author{%
  Shivam Kumar \\
  \textit{Independent Researcher} \\
  \texttt{shivam@shivamk3r.com} \\
  \url{https://shivamk3r.com/}
}

\date{May 12, 2026}

\begin{document}

\maketitle

\begin{abstract}
We introduce \textsc{ShapeCodeBench}, a synthetic benchmark for \emph{perception-to-program
reconstruction}: given a rendered raster, a model must emit an executable drawing
program that a deterministic evaluator re-renders and compares. The DSL has four
primitives on a $512 \times 512$ black-on-white canvas, but every instance is
generated from a seeded RNG, so fresh held-out sets can be minted to mitigate
benchmark contamination. Because both instance generation and scoring are
automatic, the same loop can refresh evaluations quickly without per-instance
human annotation or manual judging. We release a frozen split,
\textsc{eval\_v1} ($150$ samples, $50$ per difficulty tier), scored by exact
match, pixel accuracy, and foreground IoU alongside parse and execution
success. Evaluating four
reasoning-effort configurations of two frontier multimodal models -- \textsc{Claude
Opus 4.7} (1M context) at \texttt{high} and \texttt{max} effort, and
\textsc{GPT-5.5} at \texttt{medium} and \texttt{extra\_high} reasoning effort --
against an empty-program floor and a classical-CV heuristic baseline exposes a
tier-dependent crossover: the heuristic leads easy-tier exact match ($0.26$ vs.\
at most $0.08$ for any multimodal configuration) by individuating separated
connected components, but collapses on medium and hard scenes as overlapping
shapes fuse; the strongest multimodal model by foreground IoU (\textsc{GPT-5.5}
at \texttt{extra\_high} effort) retains most of the spatial structure and leads
foreground IoU on every tier (up to $0.87$), yet misses exact match by small
parameter errors, while \textsc{Claude Opus 4.7} (1M) trails the heuristic on
foreground IoU at both effort levels. Best overall exact match is $0.087$
(heuristic) and $0.027$ among multimodal models, so
\textsc{ShapeCodeBench} is far from saturated. Benchmark code, frozen dataset, and full
run artifacts are released to support independent replication and extension.

\end{abstract}

\section{Introduction}
\label{sec:intro}

Modern multimodal models are increasingly evaluated on their ability to turn
images into code. Work on screenshot-to-HTML~\cite{beltramelli2017pix2code,si2024design2code},
structure extraction from webpages, LaTeX, and music
scores~\cite{roberts2024image2struct}, and symbolic vector
generation~\cite{lin2025vcode,zhou2026omni} all ask the same underlying question
in different clothes: \emph{can a model look at a picture and produce the program
that generated it?} Earlier work on visual program
induction~\cite{sharma2018csgnet,liu2019scenes,li2020perspective,li2020multiplane,duan2022parametric,grand2024lilo}
framed this problem as inverse graphics over a small symbolic DSL, and more
recently \textsc{TurtleBench}~\cite{rismanchian2025turtlebench} has revived it
as a benchmark-first target for large vision-language models.

Across these lines of work, three design pressures keep recurring: (1) deterministic
execution so that scoring is principled; (2) render-based scoring so that
semantically equivalent programs are not penalized for textual
differences~\cite{roberts2024image2struct}; and (3) controlled generation so that
failure modes can be attributed to specific visual factors, in the tradition of
\textsc{CLEVR}~\cite{johnson2017clevr,bahdanau2019closure}. Most existing
benchmarks satisfy one or two of these, but few satisfy all three while remaining
\emph{renewable} -- that is, cheap enough to regenerate when an existing split
becomes contaminated. For model development, renewability also changes the
feedback cycle: a researcher can generate fresh instances, run a model, and
obtain objective scores without commissioning new labels or manual judgments for
each refreshed example.

We present \textsc{ShapeCodeBench}, a perception-to-program benchmark that attempts to
hit the intersection of these pressures. The task is narrow by design: the DSL
has exactly four primitives (\texttt{filled\_circle}, \texttt{circle},
\texttt{filled\_square}, \texttt{square}) and the canvas is fixed at
$512{\times}512$ grayscale with a black foreground on a white background. Every
sample is generated from a seeded RNG with explicit difficulty controls on shape
count, size, stroke width, overlap, and canvas clipping. The evaluator parses
predictions with a safe restricted Python parser, re-renders them through the
same deterministic Pillow pipeline used to produce the target, and compares
rasters.

\paragraph{Contributions.} We make the following contributions:
\begin{enumerate}
  \item We release the \textsc{ShapeCodeBench} benchmark: a four-primitive drawing DSL,
    a safe restricted parser, a seeded scene generator with three difficulty
    tiers, and a render-based evaluator with five primary metrics.
  \item We freeze an evaluation split, \textsc{eval\_v1}, of 150 samples with
    deterministic seeds, and publish per-sample raster hashes to make the exact
    evaluation instances reproducible across platforms.
  \item We make benchmark refresh a first-class workflow: new held-out splits can
    be generated from fresh seeds and scored automatically, enabling fast
    regression-style feedback while avoiding per-instance human annotation or
    manual judging. This mitigates exact-instance contamination but does not
    prevent models from learning the generator distribution.
  \item We release a provider-agnostic runner that records prompts, model
    configuration, raw outputs, normalized predictions, metrics, and per-sample
    artifacts, making model evaluations auditable and easy to extend.
  \item We report baseline results for four multimodal model configurations
    -- \textsc{Claude Opus 4.7} (1M context) at \texttt{high} and \texttt{max}
    effort, and \textsc{GPT-5.5} at \texttt{medium} and \texttt{extra\_high}
    reasoning effort -- alongside two non-LLM baselines (empty program,
    classical-CV heuristic). The strongest multimodal model by foreground IoU
    (\textsc{GPT-5.5}/\texttt{extra\_high}) reaches mean foreground IoU $0.87$,
    the best multimodal exact match remains $0.027$, and the classical
    heuristic leads easy-tier exact match at $0.26$ -- a
    tier-dependent crossover that confirms \textsc{ShapeCodeBench} is not saturated
    and exposes distinct failure modes in perception versus structured code
    emission. Effort tier helps modestly (\texttt{max}\,$>$\,\texttt{high} for
    Claude on FG-IoU; \texttt{extra\_high}\,$>$\,\texttt{medium} for GPT-5.5)
    but does not close either gap.
\end{enumerate}

The rest of the paper is organized as follows. Section~\ref{sec:related} places
\textsc{ShapeCodeBench} against prior work. Section~\ref{sec:benchmark} describes the
DSL, generator, renderer, and evaluator. Section~\ref{sec:experiments} details
the experimental setup and headline results. Section~\ref{sec:analysis} analyzes
failure modes, the heuristic-vs-LLM gap, and difficulty validity.
Section~\ref{sec:limitations} discusses limitations and future work.

\section{Related Work}
\label{sec:related}

\textsc{ShapeCodeBench} sits at the intersection of three established lines of work:
visual program induction, synthetic diagnostic benchmarks, and image-to-code
evaluation of multimodal models.

\paragraph{Visual program induction and inverse graphics.}
Predicting executable programs from images has a long history. CSGNet~\cite{sharma2018csgnet}
infers constructive solid geometry programs from 2D and 3D shapes, and is the
direct conceptual ancestor of \textsc{ShapeCodeBench}. \cite{liu2019scenes} learn to
describe scenes with a DSL that supports loops and grouping, demonstrating that
compositional program structure -- and not only local shape identity -- can be
recovered from a single image. \cite{li2020perspective} and
\cite{li2020multiplane} extend program induction to perspective scenes and
repeated 3D structure, while \cite{duan2022parametric} studies parametric
primitives with explicit function correlations, close in spirit to the
integer-parameter DSL of \textsc{ShapeCodeBench}. LILO~\cite{grand2024lilo} synthesizes
reusable program libraries across domains including graphics composition,
providing a template for symbolic baselines. These systems prove that the
image-to-program problem is well-defined; none of them are benchmark-first
evaluations of modern multimodal models.

\paragraph{Benchmark design.}
\textsc{CLEVR}~\cite{johnson2017clevr} showed how much scientific value a
synthetic, carefully factorized benchmark can add when it is designed to reduce
spurious shortcuts and expose reasoning failure modes.
\textsc{CLOSURE}~\cite{bahdanau2019closure} extended this insight by probing
systematic generalization. We adopt the same diagnostic posture: rather than
building a large, noisy dataset, we expose the axes of variation explicitly and
allow researchers to regenerate the dataset from a seed.

\paragraph{Renewable and dynamic evaluation.}
Renewability is becoming a first-class benchmark-design goal. \textsc{LiveBench}~\cite{white2024livebench}
adds and updates automatically scored questions to reduce test-set
contamination and avoid the failure modes of human crowdsourcing or LLM-as-judge
scoring on hard tasks. \textsc{Image2Struct}~\cite{roberts2024image2struct}
similarly emphasizes fully automatic, renewable round-trip evaluation. \textsc{ShapeCodeBench}
inherits this philosophy in a synthetic inverse-graphics setting: instead of
downloading fresh natural data, it mints new controlled instances from fresh
seeds and scores them through deterministic rendering.

\paragraph{Closest benchmark predecessors.}
\textsc{TurtleBench}~\cite{rismanchian2025turtlebench} evaluates vision-language
models on turtle-geometry programs and is the closest benchmark-level neighbor.
It reports that strong models still struggle, which supports the general thesis
that visual program reconstruction is hard. \textsc{ShapeCodeBench} differs along three
axes: (1) the DSL is a tiny shape-primitive set rather than turtle paths, which
removes path-planning reasoning and isolates perception-plus-emission; (2) the
benchmark is explicitly renewable via fresh seeds; and (3) scoring is raster-based
and deterministic.

\textsc{Image2Struct}~\cite{roberts2024image2struct} introduced round-trip
structure extraction: image $\rightarrow$ structure $\rightarrow$ rendered image
$\rightarrow$ similarity. Our scoring pipeline follows the same philosophy.
Unlike \textsc{Image2Struct}, which spans noisy real-world web pages, LaTeX, and
music, \textsc{ShapeCodeBench} stays inside a small controlled space so that reported
failures can be cleanly attributed to perception or emission.

\paragraph{Broader image-to-code.}
pix2code~\cite{beltramelli2017pix2code}, Design2Code~\cite{si2024design2code},
\textsc{VCode}~\cite{lin2025vcode}, and \textsc{Omni-I2C}~\cite{zhou2026omni}
demonstrate that image-to-code is now a mature benchmark family spanning GUI
screenshots, SVG, and general graphics-to-code. These benchmarks mix many
confounders: OCR, library conventions, rendering-engine variability, and external
assets. \textsc{ShapeCodeBench} strips away these confounders by design, providing a
sibling benchmark whose strengths are control and reproducibility rather than
realism.

\paragraph{Verifiable feedback for training.}
Automatic execution feedback has also become an important training signal for
code and reasoning models. \textsc{CodeRL}~\cite{le2022coderl} and
\textsc{RLTF}~\cite{liu2023rltf} use unit-test or execution feedback to improve
code generation, while \textsc{DeepSeek-R1}~\cite{guo2025deepseekr1} and
\textsc{RLVE}~\cite{zeng2025rlve} illustrate the broader role of verifiable
rewards and procedurally generated environments in reinforcement learning for
language models. These results motivate a future use of \textsc{ShapeCodeBench} as a
verifiable training environment, but our present contribution is an evaluation
benchmark: we do not train or fine-tune models on the task.

\paragraph{How to read the contribution.}
\textsc{ShapeCodeBench} is not the first benchmark to ask models to reconstruct
programs from images. Its contribution is a \emph{specific combination}:
a deterministic render-based evaluator, explicit difficulty axes, a
provider-agnostic adapter for cost-controlled runs, and a renewable frozen
evaluation split with publicly verifiable instance hashes. We view it as
complementary to \textsc{TurtleBench} and \textsc{Image2Struct} rather than a
replacement.

\section{Benchmark Design}
\label{sec:benchmark}

\textsc{ShapeCodeBench} is specified by four coupled components: the DSL and its
restricted parser, the scene generator, the deterministic renderer, and the
render-based evaluator. We describe each in turn.

\subsection{The ShapeCodeBench DSL}
\label{sec:dsl}

A ShapeCodeBench program is a sequence of top-level function calls, one per line.
There are exactly four primitive functions:

\begin{lstlisting}
filled_circle(cx=<int>, cy=<int>, radius=<int>)
circle(cx=<int>, cy=<int>, radius=<int>, stroke=<int>)
filled_square(cx=<int>, cy=<int>, size=<int>)
square(cx=<int>, cy=<int>, size=<int>, stroke=<int>)
\end{lstlisting}

The parser is implemented on top of Python's \texttt{ast} module but enforces a
strict whitelist: only top-level expression statements; only calls to the four
whitelisted names; only keyword arguments; only integer literals (including
\texttt{+n}/\texttt{-n} unary forms). Imports, variables, loops, comprehensions,
attribute access, starred arguments, duplicate keywords, and unexpected keywords
are rejected with typed errors. Parameter ranges are validated:
$\texttt{cx},\texttt{cy} \in [0, 511]$; $\texttt{radius},\texttt{size} \in [1, 512]$;
$\texttt{stroke} \in [1, \texttt{radius}]$ for circles and
$\texttt{stroke} \in [1, \lceil \texttt{size}/2 \rceil]$ for squares. Shapes may
extend beyond the canvas and are clipped deterministically at render time.

The serializer is canonical: one call per line, fixed keyword order per
primitive, normalized whitespace, no imports or boilerplate.

\subsection{Renderer}
\label{sec:renderer}

The renderer uses Pillow's \texttt{ImageDraw} to produce a $512 \times 512$
8-bit grayscale image. Backgrounds are white ($255$) and shapes are black ($0$).
\texttt{FilledCircle} uses \texttt{draw.ellipse(fill)}; \texttt{Circle} uses
\texttt{draw.ellipse(outline, width=stroke)}; and the square counterparts use
\texttt{draw.rectangle} with the same conventions. Program order is preserved
in the render loop, but the binary palette makes scenes order-invariant: later
shapes can add foreground pixels but cannot erase them. True order-sensitive
evaluation is deferred to a future version.

\subsection{Generator and Difficulty Tiers}
\label{sec:generator}

The generator is seeded by a single integer and uses only
\texttt{random.Random(seed)} for determinism. Scenes are produced by
rejection-sampling candidate shapes until they satisfy tier-specific constraints
on (i) shape count, (ii) primitive extent (radius or size), (iii) stroke width,
(iv) canvas clipping probability, and (v) the maximum allowed bounding-box IoU
between the new shape and any existing shape. A tier may additionally require at
least one pairwise bounding-box overlap.

\begin{table}[t]
  \centering
  \small
  \begin{tabular}{lcccccc}
  \toprule
  Tier & \# shapes & Extent & Stroke & Clip prob. & Max bbox IoU & Overlap? \\
  \midrule
  Easy & $1$--$3$ & $[64, 160]$ & $[2, 6]$ & $0.00$ & $0.02$ & no \\
  Medium & $3$--$6$ & $[32, 128]$ & $[2, 8]$ & $0.25$ & $0.35$ & no \\
  Hard & $6$--$10$ & $[16, 128]$ & $[1, 10]$ & $1.00$ & $-$ & yes \\
  \bottomrule
  \end{tabular}
  \caption{Difficulty tiers used to structure generation in \textsc{ShapeCodeBench}. ``Extent'' is the radius or size range;
  a missing Max bbox IoU indicates no constraint.}
  \label{tab:difficulty-tiers}
\end{table}

Each generated sample is written to disk as a PNG together with a JSON metadata
file containing the sample ID, split, difficulty, seed, canvas size, shape count,
shape inventory, ground-truth program, and render configuration. Our frozen
evaluation split \textsc{eval\_v1} uses contiguous seeds $0{-}49$ per tier,
yielding 150 samples total; their SHA-256 hashes are published alongside the
dataset.

\begin{figure}[t]
  \centering
  \includegraphics[width=0.85\linewidth]{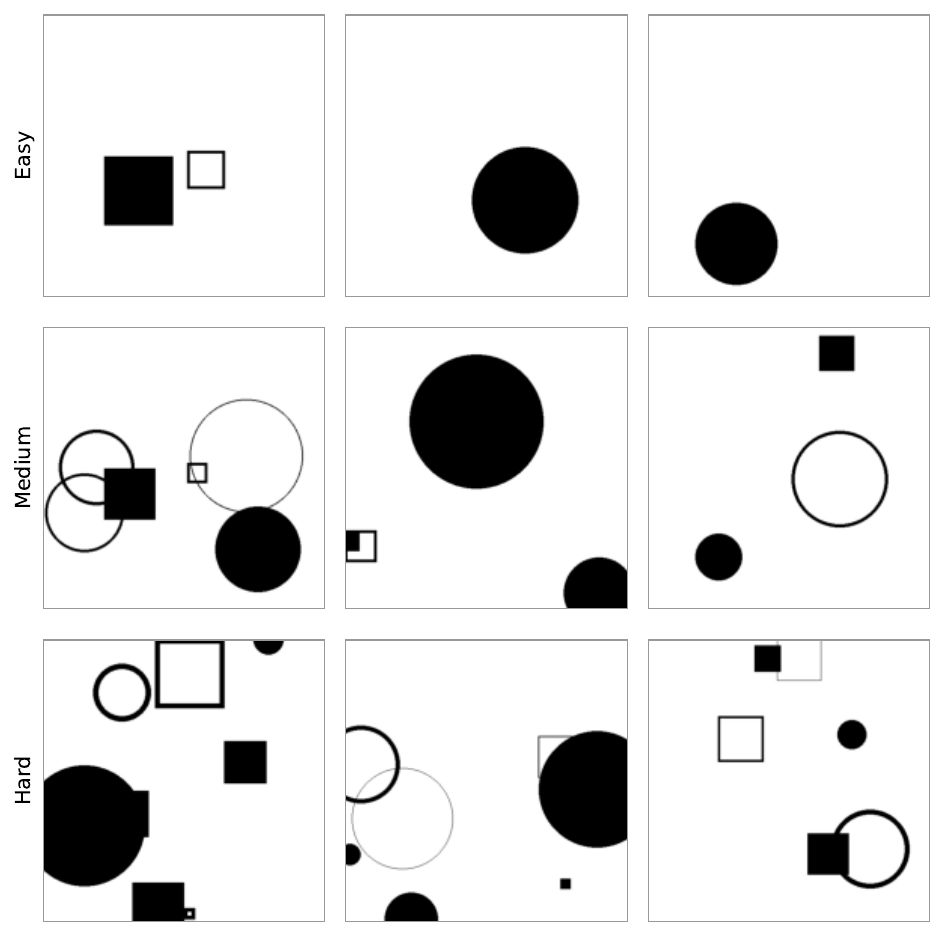}
  \caption{Representative samples from \textsc{eval\_v1} at each difficulty
  tier. The three tiers differ in shape count, extent, and overlap.}
  \label{fig:sample-grid}
\end{figure}

\subsection{Evaluator}
\label{sec:evaluator}

Given a target image $I_t$ and a predicted DSL program $p$, the evaluator
attempts to parse $p$ through the restricted parser, render the resulting scene
into $I_p$, and compare rasters. We report five metrics.

\begin{itemize}
  \item \textbf{Exact match} -- $1$ if $I_t = I_p$ pixel-exactly, else $0$.
  \item \textbf{Pixel accuracy} -- fraction of pixels equal between $I_t$ and $I_p$.
  \item \textbf{Foreground IoU} -- intersection-over-union of black pixels
    between $I_t$ and $I_p$ (convention: if both sets are empty, IoU is $1$).
  \item \textbf{Parse success} -- $1$ if the parser accepts $p$, else $0$.
  \item \textbf{Execution success} -- $1$ if rendering the parsed scene succeeds, else $0$.
\end{itemize}

On parse or execution failure, all similarity metrics fall to $0$ and the
failure type is recorded for later analysis. We aggregate metrics over the full
split and also report per-tier breakdowns.

\section{Experiments}
\label{sec:experiments}

\subsection{Protocol}

We evaluate six systems on the frozen \textsc{eval\_v1} split: two non-LLM
baselines and four reasoning-effort configurations across two frontier
multimodal models. Exact invocation details are provided in
Appendix~\ref{app:reproducibility}.
\begin{itemize}
  \item \textbf{Empty-Program} -- a floor baseline that always predicts an empty
    string; every sample fails parsing.
  \item \textbf{Heuristic-CV} -- a classical-CV baseline that thresholds the
    image, labels connected components, classifies each component as a circle or
    square by bounding-box fill ratio, and as hollow or filled by morphological
    erosion. Stroke widths are estimated from the ratio of component area to
    estimated perimeter.
  \item \textbf{\textsc{Claude Opus 4.7} (1M context)} at \texttt{high} and
    \texttt{max} effort.
  \item \textbf{\textsc{GPT-5.5}} at \texttt{medium} and \texttt{extra\_high}
    reasoning effort.
\end{itemize}

All four LLM configurations share the same zero-shot prompt: a one-sentence
system instruction (``Return only valid ShapeCodeBench DSL code. Do not include
markdown fences, comments, or prose.'') and a user block listing the four
primitive signatures and formatting constraints. We do not use chain-of-thought
prompting or few-shot examples. Raw model outputs are post-processed by a shared
normalizer that prefers fenced code blocks anywhere in the response, falls back
to primitive-signature line filtering, and ultimately surfaces the raw response
so that parse errors are visible rather than masked.

\subsection{Runner and artifacts}

Each run writes a per-run directory under \texttt{data/runs/} containing
\texttt{run\_config.json}, \texttt{summary.json}, and per-sample JSON files with
the request, raw and normalized predictions, usage, latency, and the full
evaluation result. All metrics in this paper are computed from these artifacts by
\texttt{scripts/analyze.py}; figures are produced by
\texttt{scripts/make\_figures.py}. We use non-parametric bootstrap with 1000
resamples for 95\% confidence intervals on per-difficulty means.

\subsection{Headline results}

\begin{figure}[!htbp]
  \centering
  \includegraphics[width=\linewidth]{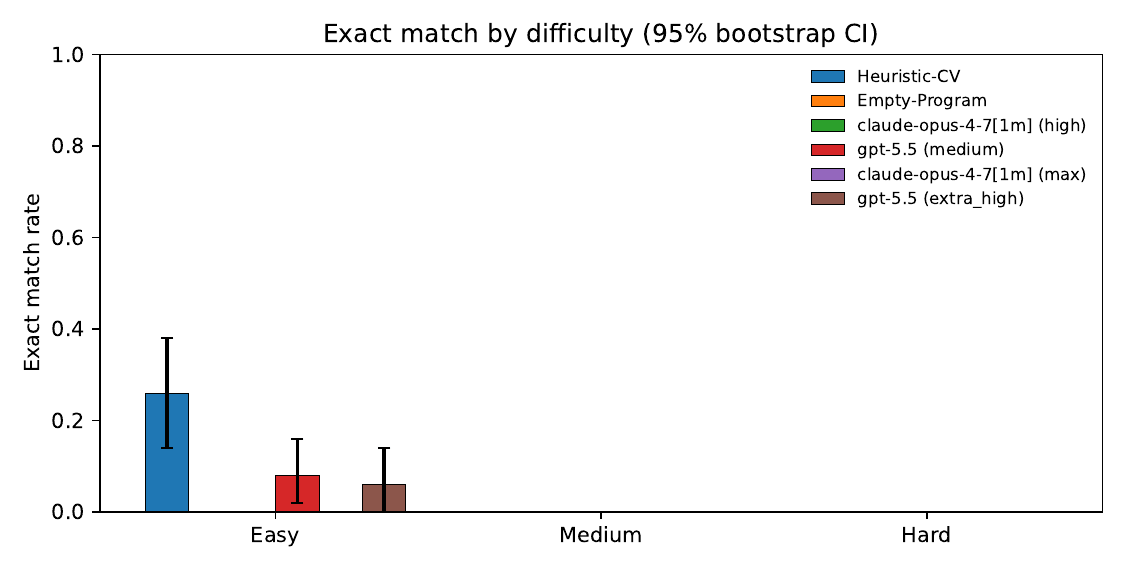}
  \caption{Exact-match rate by difficulty tier. Error bars are 95\% bootstrap
  CIs over per-sample indicators.}
  \label{fig:acc-by-diff}
\end{figure}

\begin{figure}[!htbp]
  \centering
  \includegraphics[width=\linewidth]{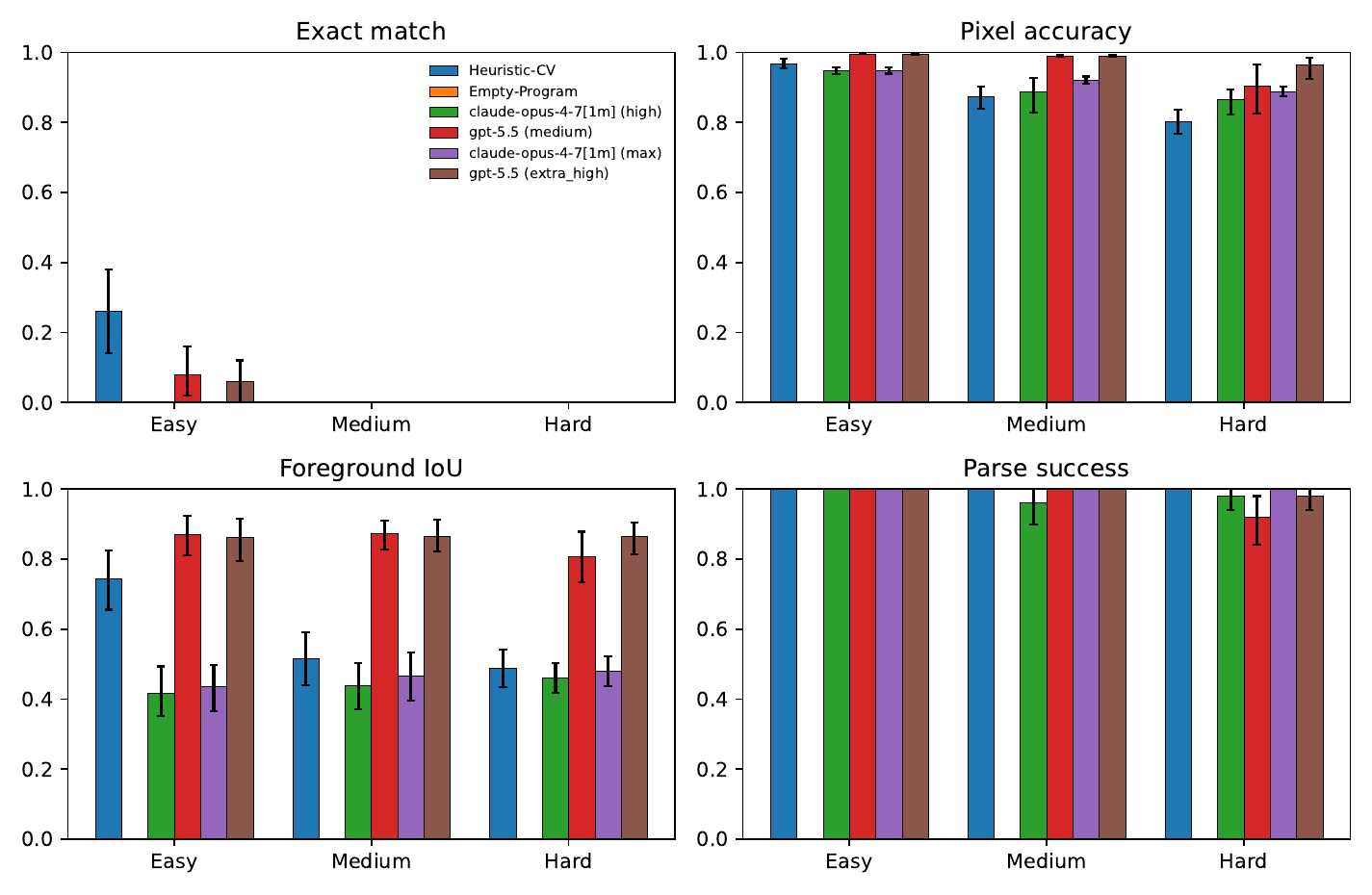}
  \caption{All four scored metrics by difficulty tier. The four panels are
  exact-match, pixel accuracy, foreground IoU, and parse success.}
  \label{fig:metric-panel}
\end{figure}

Table~\ref{tab:main-results} reports the aggregated metrics across all 150
samples for each system; 95\% bootstrap CIs are shown in brackets.
Figure~\ref{fig:acc-by-diff} breaks the exact-match rate down by difficulty
tier, and Figure~\ref{fig:metric-panel} shows the same decomposition for all
four scored metrics.

\begin{table}[!htbp]
  \centering
  \small
\begin{tabular}{lrrrr}
\toprule
Model & Exact & PixAcc & FG-IoU & Parse \\
\midrule
Heuristic-CV & 0.087 {\scriptsize [0.047, 0.133]} & 0.881 {\scriptsize [0.862, 0.898]} & 0.583 {\scriptsize [0.535, 0.628]} & 1.000 {\scriptsize [1.000, 1.000]} \\
gpt-5.5 (medium) & 0.027 {\scriptsize [0.007, 0.053]} & 0.963 {\scriptsize [0.937, 0.984]} & 0.850 {\scriptsize [0.813, 0.884]} & 0.973 {\scriptsize [0.947, 0.993]} \\
gpt-5.5 (extra\_high) & 0.020 {\scriptsize [0.000, 0.040]} & 0.983 {\scriptsize [0.969, 0.991]} & 0.865 {\scriptsize [0.836, 0.893]} & 0.993 {\scriptsize [0.980, 1.000]} \\
Empty-Program & 0.000 {\scriptsize [0.000, 0.000]} & 0.000 {\scriptsize [0.000, 0.000]} & 0.000 {\scriptsize [0.000, 0.000]} & 0.000 {\scriptsize [0.000, 0.000]} \\
claude-opus-4-7[1m] (high) & 0.000 {\scriptsize [0.000, 0.000]} & 0.900 {\scriptsize [0.873, 0.919]} & 0.439 {\scriptsize [0.403, 0.474]} & 0.980 {\scriptsize [0.960, 1.000]} \\
claude-opus-4-7[1m] (max) & 0.000 {\scriptsize [0.000, 0.000]} & 0.919 {\scriptsize [0.912, 0.927]} & 0.461 {\scriptsize [0.427, 0.495]} & 1.000 {\scriptsize [1.000, 1.000]} \\
\bottomrule
\end{tabular}

  \caption{Main results on \textsc{eval\_v1}. ``Exact'' is pixel-exact raster
  match; ``PixAcc'' is mean pixel accuracy; ``FG-IoU'' is mean foreground IoU;
  ``Parse'' is parse success rate. Brackets give 95\% bootstrap CIs.}
  \label{tab:main-results}
\end{table}

The key qualitative pattern -- detailed in Section~\ref{sec:analysis} -- is that
exact-match collapses on the hard tier for every system, while
foreground IoU degrades more gracefully; the heuristic baseline is surprisingly
competitive on easy scenes and is outclassed on hard scenes, where LLMs can
enumerate and place multiple overlapping shapes that classical connected
components cannot individuate.

\section{Analysis}
\label{sec:analysis}

\subsection{Error taxonomy}

\begin{figure}[!htbp]
  \centering
  \includegraphics[width=\linewidth]{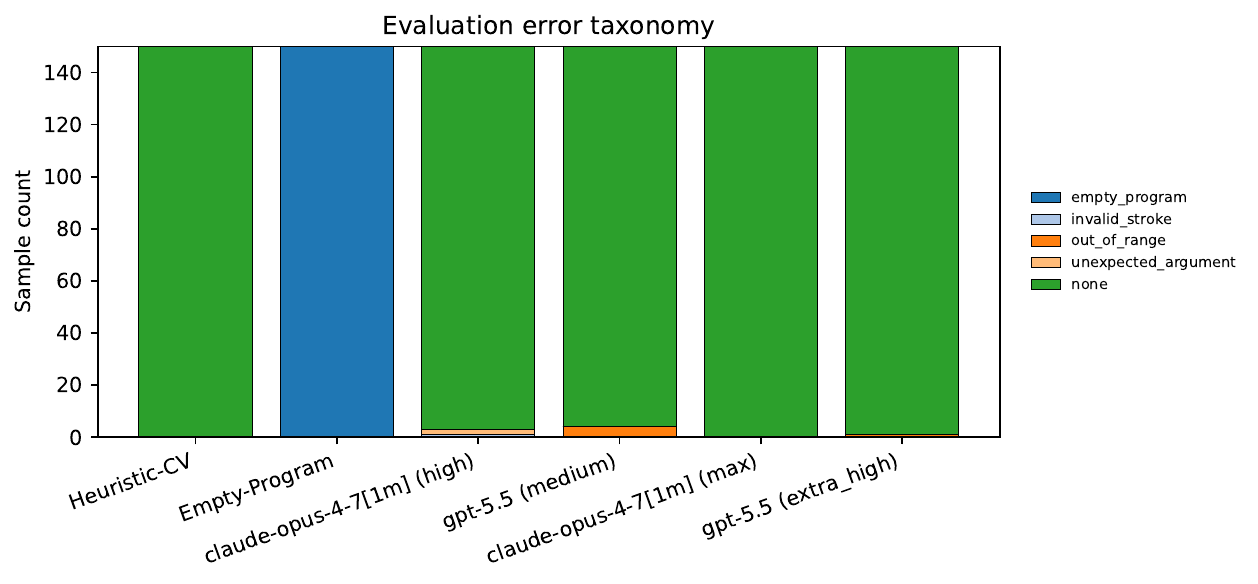}
  \caption{Evaluator error taxonomy per model. The ``none'' bar counts samples
  whose predictions parsed and rendered without error, including samples that
  did not pixel-match the target. Remaining categories surface structured
  parsing or validation failures.}
  \label{fig:error-hist}
\end{figure}

Figure~\ref{fig:error-hist} reports the distribution of evaluator error types
per model. Three patterns are worth flagging.

First, \texttt{Empty-Program} concentrates all 150 samples under the
\texttt{empty\_program} parse error, because an empty DSL program is a parse
failure by construction. This is the intended floor and not a pathology.

Second, the LLM runs have non-zero but small parse-failure counts, dominated by
\texttt{out\_of\_range} (predicted coordinates or extents outside the valid
$[0,511]$ and $[1, 512]$ ranges) and \texttt{invalid\_stroke} (stroke widths
exceeding the primitive's documented limit). These are cases where the model
understood the task format but violated structural constraints -- the kind of
failure mode that shows models do not internalize the DSL's range constraints
from a short prompt alone.

Third, the \texttt{Heuristic-CV} baseline has \emph{zero} parse failures because
it only emits programs it can itself construct. Its errors manifest as low
foreground IoU rather than parse failures.

\subsection{Qualitative wins and losses}

\begin{figure}[!htbp]
  \centering
  \includegraphics[width=0.95\linewidth]{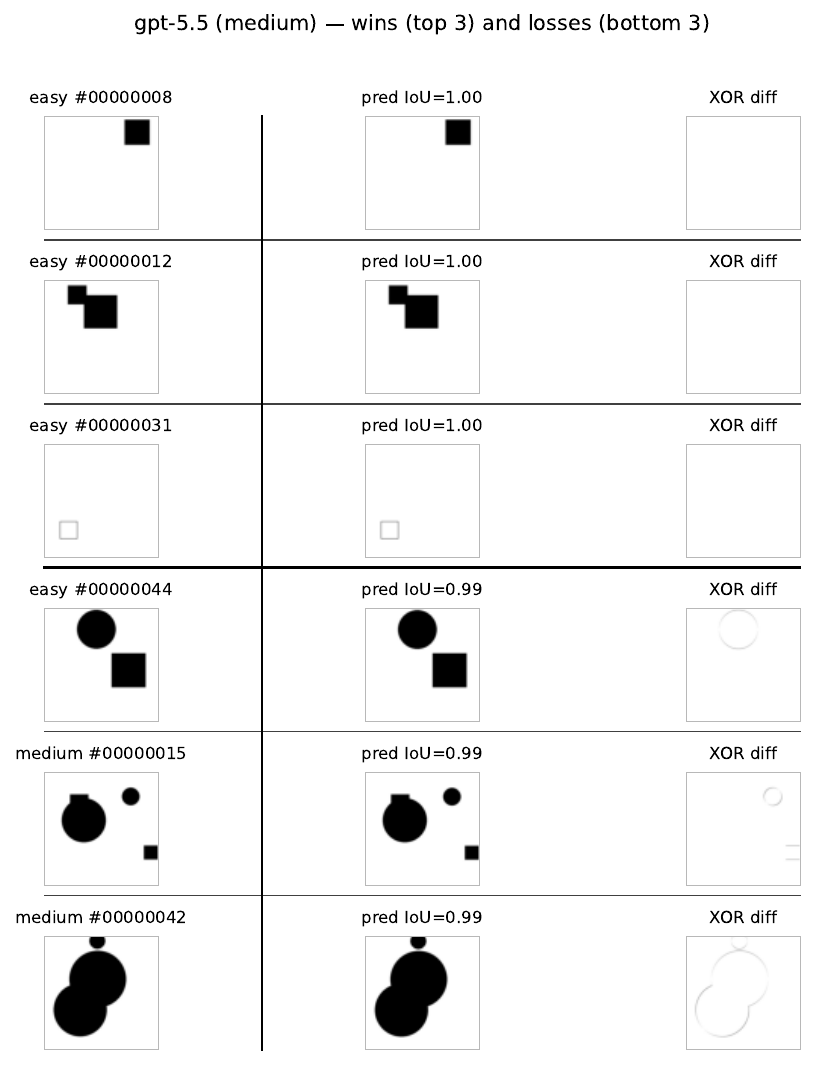}
  \caption{Wins (top) and losses (bottom) for the best exact-match multimodal
  configuration on \textsc{eval\_v1} (\textsc{GPT-5.5} at \texttt{medium}
  reasoning effort, selected automatically by exact-match rate). Each row shows
  the target, the re-rendered prediction, and a foreground-XOR diff. Losses
  cluster on overlapping hard scenes and stroke-width confusions.}
  \label{fig:qualitative}
\end{figure}

Figure~\ref{fig:qualitative} shows representative wins (top rows) and losses
(bottom rows) for the best exact-match multimodal configuration: each row shows
the target image, the prediction rendered through the same renderer, and the XOR
diff of foreground masks. The wins are dominated by the easy tier, where two or
three
non-overlapping shapes can be located and parameterized precisely. Losses tend
to come from medium and hard tiers and split into three recurring categories:
(i) correct shape inventory but off-by-a-few-pixel parameter estimates, (ii)
missed occluded shapes in high-overlap hard scenes, and (iii) misclassification
of hollow vs. filled when stroke widths are thin.

\subsection{Difficulty validity}

A renewable benchmark is useful only if its difficulty tiers track real
performance gradients. Figure~\ref{fig:acc-by-diff} shows that exact-match rate
falls monotonically from easy to hard for every system, including the heuristic
baseline. Foreground IoU tracks the same ordering but with shallower slopes, as
expected: pixel-level similarity degrades more gracefully than the all-or-nothing
exact-match metric. This monotonic structure supports the claim that
\textsc{eval\_v1}'s difficulty axes are not arbitrary.

\subsection{Heuristic vs. LLM gap}

The heuristic baseline anchors what a purely bottom-up computer-vision pipeline
with no DSL-level reasoning can achieve. Its story splits by tier. On the easy
tier it is surprisingly competitive on \emph{exact-match}, because easy scenes
are separated and unclipped -- connected components match shapes directly, the
area/perimeter stroke estimator is close enough, and the hollow-vs-filled test
on eroded masks rarely errs. Multimodal models, by contrast, almost always
miss exact-match on easy scenes by a few pixels of parameter error: they
recover the \emph{right shapes} but not the \emph{right parameter values}.

On the medium and hard tiers the picture flips. Foreground IoU for the
heuristic deteriorates sharply because overlapping or clipped shapes merge
into a single connected component and the pipeline can no longer individuate
them. Multimodal models retain most of the spatial structure (foreground IoU
stays roughly flat across tiers) but still do not pixel-match -- their
programs contain the right number of shapes in roughly the right positions,
but they cannot land parameters precisely under occlusion.

Taken together, the heuristic is a better \emph{exact-match} baseline on easy
scenes and a worse \emph{IoU} baseline on harder ones. This decomposition is
precisely the kind of diagnostic signal \textsc{ShapeCodeBench} is designed to
produce: difficulty is not a single-axis phenomenon, and different systems
fail in different places. The sharp easy-tier exact-match comparison also
argues that closing that gap -- having a model emit \emph{both} the right
shape list \emph{and} the right parameter values on clean scenes -- is a
natural first-order target for future work.

\section{Limitations and Future Work}
\label{sec:limitations}

\textsc{ShapeCodeBench} makes deliberate scoping choices in its v1 incarnation, each
of which is a direction for future work.

\paragraph{Monochrome palette.}
V1 is black-on-white. This collapses draw-order sensitivity: later shapes can
paint foreground pixels but cannot erase or overwrite earlier ones. A richer
palette (multiple colors or an explicit \texttt{clear} primitive) would make
draw order first-class and enable sharper compositionality tests.

\paragraph{Four primitives.}
The DSL currently covers only filled and hollow circles and squares. Extending
to rectangles, lines, polygons, or parametric curves would stress different
kinds of visual reasoning and likely move saturation further out.

\paragraph{Zero-shot only.}
We evaluate without chain-of-thought prompting or few-shot examples. These are
natural knobs to explore and may change the ordering of models, especially for
the reasoning-heavy hard tier.

\paragraph{Model-inference variability.}
The frozen evaluation images, parser, renderer, and scorer are deterministic:
regenerating \textsc{eval\_v1} from the published seeds should reproduce the
same target PNGs and metric computation. The remaining variability is in model
inference. Repeating the same image-and-prompt request to a hosted multimodal
model may produce a different predicted program, so we cannot guarantee
bit-exact reproduction of the reported model scores. Each run's
\texttt{run\_config.json} records the model configuration and invocation settings
needed to reproduce the experimental setup.

\paragraph{No human baseline.}
We do not report human performance. An informal human baseline on a small
sample of scenes would make the ``how hard is this really?'' claim concrete,
and we plan to add it in a subsequent revision.

\paragraph{Model coverage.}
We evaluate two frontier multimodal models -- \textsc{Claude Opus 4.7} (1M
context) and \textsc{GPT-5.5} -- across two reasoning-effort tiers each. Other
frontier multimodal systems are not covered here but can be added by
implementing a new adapter against the same \texttt{ModelAdapter} Protocol.

\paragraph{Contamination resistance.}
The current eval split's seeds are public. We rely on the \emph{renewability}
of the benchmark rather than on secrecy: to evaluate under a contamination-free
setting, regenerate \textsc{eval\_vN} from fresh seeds using the published
generator and manifest. This is intentionally cheap: the entire 150-sample split
regenerates in under a second.

\paragraph{Training signal, not a training pipeline.}
The current evaluator is an offline benchmark, not a differentiable pretraining
loss. Predictions are discrete DSL text, the parser is symbolic, the renderer is
Pillow-based, and the reported metrics are computed after hard rasterization.
Future training use would therefore require a separate setup, such as supervised
fine-tuning on generated image/program pairs, reinforcement learning or
rejection sampling with render-based rewards, a learned reward or critic model,
or a differentiable renderer variant. Any such use should train on generated
training seeds and reserve \textsc{eval\_v1} or fresh held-out splits for clean
evaluation.

\paragraph{Reproducibility.}
We publish the benchmark code, the frozen \textsc{eval\_v1} dataset with
SHA-256 hashes per image, per-run artifacts, analysis scripts, and paper sources.
\texttt{docs/REPRODUCIBILITY.md} walks through the end-to-end workflow from a
clean checkout.

\section*{Acknowledgments}

We thank the authors of the prior benchmarks cited throughout this paper for
laying the groundwork that \textsc{ShapeCodeBench} builds on, and the maintainers of
\texttt{Pillow}, \texttt{numpy}, and \texttt{scipy} for tools that made the
evaluator and heuristic baseline straightforward to implement.

\bibliographystyle{plain}
\bibliography{references}

\appendix
\section{Reproducibility Details}
\label{app:reproducibility}

The main paper reports model identifiers and reasoning-effort settings; this
appendix records the operational path used for the reported sweeps. The full
step-by-step workflow is maintained in \path{docs/REPRODUCIBILITY.md}, and each
run also writes its exact configuration to
\path{data/runs/<run_id>/run_config.json}.

\paragraph{OpenAI Codex invocation.}
\textsc{GPT-5.5} runs used the OpenAI Codex CLI, authenticated through the
author's ChatGPT subscription. The adapter invokes this command shape:
\begin{lstlisting}
codex exec --skip-git-repo-check --ephemeral -s read-only \
  -m gpt-5.5 -i <image_path> -o <last_message_path> \
  -C <temporary_workdir> --color never \
  -c reasoning_effort=<medium|extra_high>
\end{lstlisting}
The image is attached once per request, and \texttt{-o} captures the final
message for normalization. The paper sweeps used two retries with exponential
backoff and per-sample timeouts of 1800 seconds for \texttt{medium} and 2400
seconds for \texttt{extra\_high}.

\paragraph{Claude invocation.}
\textsc{Claude Opus 4.7} (1M context) runs used the Claude Code CLI,
authenticated through the author's Claude subscription. The adapter invokes this
command shape:
\begin{lstlisting}
claude --print --add-dir <image_parent> \
  --model claude-opus-4-7[1m] --effort <high|max> \
  --output-format text --no-session-persistence
\end{lstlisting}
The target image is referenced in the prompt as \texttt{@<image\_path>}, and
\texttt{--add-dir} grants read access to the sample directory. The paper sweeps
used two retries with exponential backoff and per-sample timeouts of 1800
seconds for \texttt{high} and 2400 seconds for \texttt{max}.

\paragraph{Artifacts.}
Both multimodal paths receive the same zero-shot prompt and are normalized by the
same prediction normalizer before parsing and rendering. Run artifacts include
the raw response, normalized response, latency, adapter metadata, full evaluation
result, and aggregate summaries. The non-LLM baselines use the same runner and
artifact schema but do not call an external model-serving interface.

\end{document}